\documentclass{article}
\usepackage{url}
\usepackage{amsmath,amsfonts,amssymb}
\usepackage{graphicx}
\usepackage{booktabs}
\usepackage{geometry}
\usepackage{wrapfig}
\usepackage{needspace}
\usepackage{float}
\usepackage{caption}
\newcommand{\safeimage}[2]{\IfFileExists{#1}{\includegraphics[width=#2]{#1}}{\fbox{\parbox[c][1.1in][c]{#2}{\centering Figure asset pending: \texttt{\detokenize{#1}}}}}}

\usepackage{listings}
\usepackage{xcolor}

\title{Variational Template Matching with Semivariogram-Guided Statistical Detection and Image-Adaptive Fusion}
\author{Qinwu Xu, Yifan Jiang}
\date{}

\begin{document}
\maketitle
\begin{abstract}
Reliable detection of debris and structural defects in patterned images is challenging when anomaly shapes vary and labeled examples are scarce. We develop an interpretable detection framework integrating variational normalized cross-correlation (NCC) template matching and a semivariogram-guided kernel-density estimation (KDE). Detector-specific scores can be combined through image-adaptive fusion of calibrated signed margins, without gradient-based neural-network training. We distinguish the expanded tri-modal formulation from the currently available biological-image experiments, which evaluate NCC--KDE fusion, and retain earlier geometric-matching results under their original evaluation protocol. Across training fractions, the two-detector fusion achieves F1 scores of 0.912--0.938 at 5--20\% labeled data, although these operating points exceed the specified 5\% false-positive-rate (FPR) limit. At 50\% training, fusion achieves F1=0.924 and FPR=0.039, compared with F1=0.992 and FPR=0.010 for a selected ResNet-50 checkpoint. Fusion also exceeds the CNN's F1 after one epoch across the reported fractions, illustrating early-training competitiveness rather than superiority at matched FPR. The study characterizes the benefits and limits of complementary, calibration-based detectors and identifies the controlled experiments needed to evaluate tri-modal fusion conclusively.
\end{abstract}
\section{Introduction}
Detecting small debris, breakage, and other localized defects in patterned imagery is important in biological imaging, industrial inspection, and materials characterization. In these settings, anomalies often interrupt repetitive structures while varying in size, orientation, shape, and appearance. Reliable detection is particularly difficult when annotations are expensive and the available training set is small.

Convolutional neural networks (CNNs) can learn discriminative representations directly from examples, but their performance depends on training data, optimization, and computational resources. In contrast, template matching provides a transparent way to identify recognizable defect structures without training a feature extractor. Conventional fixed-template matching, however, is sensitive to changes in scale, orientation, and projective geometry. A single structural similarity score may also miss defects whose appearance differs from the selected templates.

We address geometric variability by matching against a family of transformed templates using normalized cross-correlation (NCC). This structural detector is complemented by another distinct source of evidence: semivariogram-guided kernel-density estimation (KDE), which screens spatially irregular patches and evaluates their intensity distributions. Each detector produces an interpretable score; image-adaptive fusion weights their standardized signed margins to produce a final decision. Although the underlying feature computations require no neural-network optimization, detector thresholds and fusion parameters are calibrated from labeled training data.

The present manuscript consolidates earlier variational-matching experiments and newer training-fraction comparisons. The reported biological-image fusion experiments currently combine NCC and KDE. We preserve this distinction rather than attributing two-detector measurements to the expanded three-detector method.

Our empirical analysis examines two complementary questions: how reliably the detectors operate under a strict FPR target, and how their performance compares with ResNet-50 after limited versus extended training. The resulting evidence highlights both the appeal of calibrated, non-neural detection in low-data settings and the accuracy advantage achievable with sufficient CNN training.

\section{Related Work}

\subsection{Classical Template Matching}

Template matching is a fundamental technique in computer vision for detecting objects or patterns by comparing a predefined template with local image regions. Classical formulations are typically based on cross-correlation and normalized cross-correlation (NCC), which remain widely used due to their simplicity and interpretability \cite{Brunelli2009,Szeliski2022}. In particular, NCC provides robustness to linear intensity changes and has been extensively used in practical systems \cite{Lewis1995,Briechle2001}.

Despite these advantages, classical template matching relies on a fixed template and is therefore sensitive to geometric variations such as scale, rotation, and perspective distortion. This limitation becomes especially restrictive in anomaly detection tasks where defect shapes and orientations can vary significantly.

\subsection{Deformation-Robust Template Matching}

To improve robustness, several extensions to template matching have been proposed, including multi-scale representations, image pyramids, and geometric transformations such as rotation and affine mapping \cite{Szeliski2022}. These approaches allow matching across different resolutions and orientations, partially addressing the limitations of fixed-template methods.

Fast implementations of normalized cross-correlation have also been developed to improve computational efficiency in large-scale matching problems \cite{Lewis1995,Briechle2001}. However, most prior work focuses on object detection and image registration rather than anomaly detection in structured or repetitive patterns. Moreover, transformation handling is often implemented heuristically rather than formulated within a unified optimization framework.

\subsection{Statistical Anomaly Detection}

Statistical approaches detect anomalies by identifying deviations from expected data distributions. In image analysis, this often involves modeling local pixel intensity distributions using histograms, kernel density estimation, or related statistical descriptors. Such methods are particularly attractive in unsupervised settings, as they do not require labeled anomaly data \cite{Tahmasebi2012}.

However, purely statistical approaches typically lack explicit spatial and structural awareness. As a result, they may be sensitive to threshold selection and can produce high false positive rates when normal background patterns exhibit natural variability. This motivates combining statistical signals with structural methods.

\subsection{Deep Learning for Anomaly Detection}

Deep learning approaches, particularly convolutional neural networks (CNNs), have achieved strong performance in image classification and anomaly detection \cite{He2016}. The MVTec AD dataset has become a standard benchmark for industrial anomaly detection and has driven significant progress in this area \cite{Bergmann2019}.

Recent methods model feature distributions extracted from pretrained networks for anomaly detection and localization. For example, PaDiM models patch-wise feature distributions \cite{Defard2021}, while PatchCore leverages memory banks for improved recall in industrial inspection tasks \cite{Roth2022}. Other approaches, such as CutPaste, employ self-supervised learning to improve anomaly sensitivity \cite{Li2021}.

While these methods achieve strong performance, they typically require pretrained models, feature extraction pipelines, or training procedures, and may lack interpretability. In addition, many approaches focus primarily on image-level classification unless specifically designed for localization.

\section{Contributions}
Our contributions comprise (i) variational geometric template matching for localization, (ii) semivariogram-based spatial filtering followed by KDE intensity-distribution scoring, (iii) training-calibrated, image-dependent tri-modal fusion, and (iv) a comparison of detector performance across available training fractions under a strict FPR target. We separate earlier reported baseline results from the newer split-dependent evaluations because the protocols have not been established as identical.
\section{Method}
\subsection{Overview}
Given an image $I$, we extract three complementary anomaly scores: geometric template similarity ($x_T$), frequency-domain texture deviation ($x_F$), and semivariogram-guided intensity-distribution deviation ($x_K$). Each detector is configured independently before score-level fusion. The resulting framework provides image-level classification; spatial patch locations also support localization for the patch-based detectors.

\subsection{Variational Template Matching}

\subsubsection*{Correlation-Based Matching}

We use normalized cross-correlation (NCC) to measure the similarity between an image patch $I_x$ and an anomaly template $T$:

\begin{equation}
\mathrm{NCC}(I_x,T)=
\frac{
\sum_{i,j}(I_x(i,j)-\mu_{I_x})(T(i,j)-\mu_T)
}{
\sqrt{
\sum_{i,j}(I_x(i,j)-\mu_{I_x})^2
\sum_{i,j}(T(i,j)-\mu_T)^2
}
},
\end{equation}
where $\mu_{I_x}$ and $\mu_T$ are the corresponding mean intensities. NCC reduces sensitivity to intensity variations compared with unnormalized cross-correlation.

\subsubsection*{Variational Template Forms}

To accommodate geometric variations in anomaly patterns, we generate a family of templates through pyramid scaling, rotation, and perspective transformations.

\paragraph{Pyramid Scaling}
We use three scales,
$\mathcal{S}=\{1.0,0.5,0.25\}$,
to detect anomalies of different sizes.

\paragraph{Rotation and Perspective Transformation}
Templates are rotated at
$\theta\in\{0^\circ,45^\circ,90^\circ\}$.
Rotation, translation, and perspective deformation can be represented by the homogeneous transformation:

\begin{equation}
\begin{bmatrix}
u\\v\\w
\end{bmatrix}
=
\begin{bmatrix}
\cos\theta & -\sin\theta & c\\
\sin\theta & \cos\theta & f\\
P_1 & P_2 & 1
\end{bmatrix}
\begin{bmatrix}
x\\y\\1
\end{bmatrix},
\qquad
(x',y')=\left(\frac{u}{w},\frac{v}{w}\right).
\end{equation}

Here, $c$ and $f$ denote translation parameters, while $P_1$ and $P_2$ control perspective deformation. The transformation matrix is estimated from four point correspondences using OpenCV. Spatial translation during detection is handled implicitly by sliding-window matching.

\begin{wrapfigure}{l}{0.4\textwidth}
\centering
\safeimage{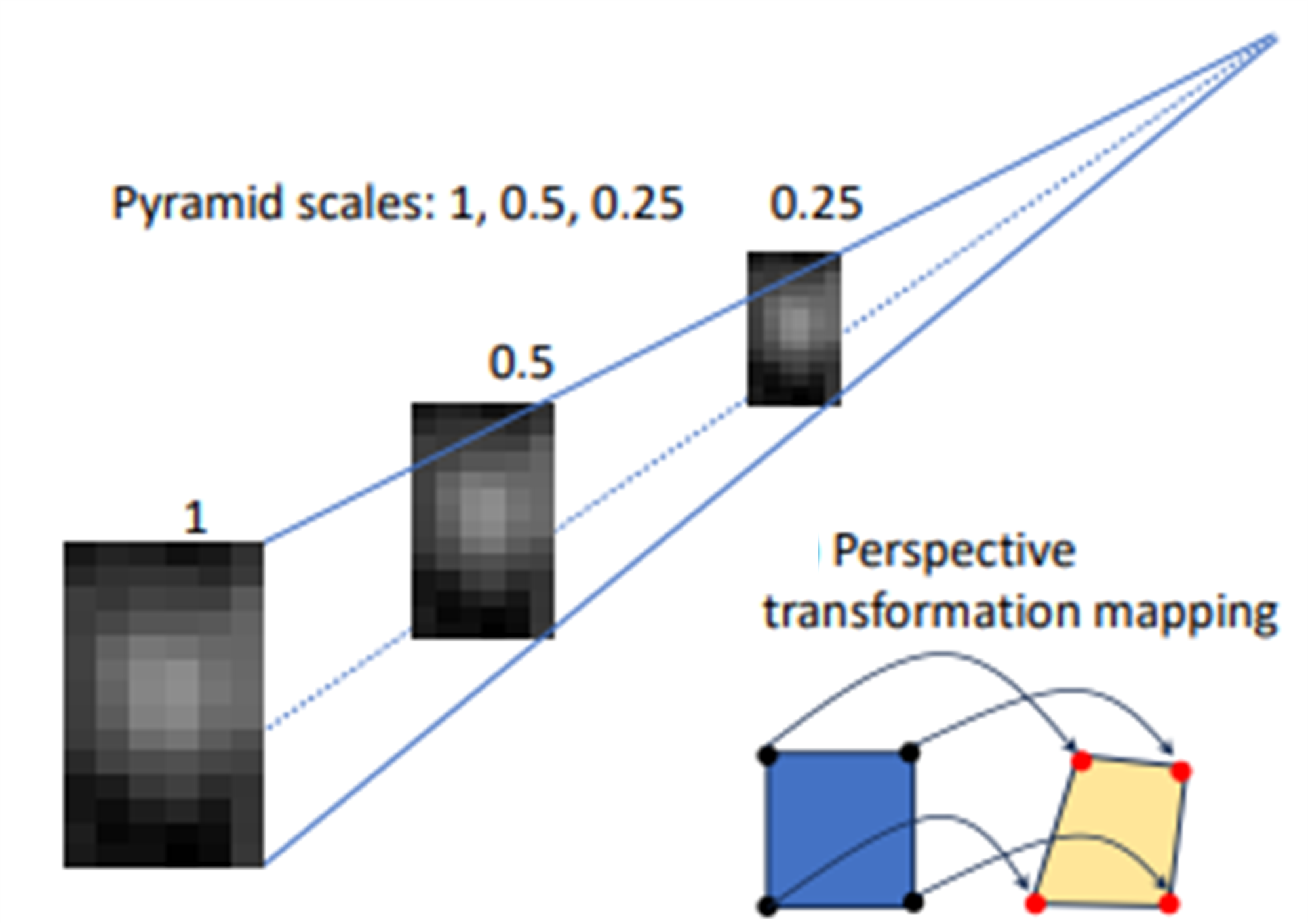}{\linewidth}
\caption{\small Variational template forms:
(1) pyramid scaling at 100\%, 50\%, and 25\%;
(2) four-point mapping for perspective transformation.}
\label{fig:fig4}
\end{wrapfigure}

\paragraph{Variational Matching Score}
Given the family of transformed templates
$\mathcal{T}=\{T_g:g\in\mathcal{G}\}$,
the matching score at location $x$ is

\begin{equation}
S_{\mathrm{tmpl}}(x)=
\frac{1+\max_{g\in\mathcal{G}}
\mathrm{NCC}(I_x,T_g)}{2}.
\end{equation}

The maximum score over spatial locations provides the image-level template score $x_{T,i}$ used in fusion. The maximizing location also supplies a candidate anomaly location. The scale and transformation parameters of the selected template describe the detected geometric variation.

\subsection{Semivariogram-Guided KDE Detector}
For a patch $P$, directional semivariances at lag $h$ are
\begin{equation}
 \gamma_{a}(P;h)=\frac{1}{2|\mathcal N_a(h)|}\sum_{(p,q)\in\mathcal N_a(h)}[P(p)-P(q)]^2,\quad a\in\{x,y\}.
\end{equation}
Spatially unusual patches are screened using vertical semivariance and the anisotropy ratio $r(P;h)=\gamma_y(P;h)/(\gamma_x(P;h)+\epsilon)$. Candidate patches satisfy a training-derived upper-quantile condition on either statistic:
\begin{equation}
 \mathcal C_i=\{P:\gamma_y(P)\ge q_\gamma\ \lor\ r(P)\ge q_r\}.
\end{equation}
For each candidate patch, a Gaussian kernel density estimate is fitted to inverted pixel intensities $v_1,\ldots,v_n$:
\begin{equation}
 \hat f_P(v)=\frac{1}{nb}\sum_{k=1}^n K\!\left(\frac{v-v_k}{b}\right).
\end{equation}
The KDE anomaly statistic measures the inverse slope of the corresponding cumulative intensity distribution over a prescribed percentile interval. The image-level score is the largest candidate-patch response, $x_{K,i}=\max_{P\in\mathcal C_i}s_{\rm KDE}(P)$. Spatial quantiles and KDE parameters are estimated from the training partition. This cascade first restricts the search spatially, then characterizes intensity-distribution irregularity.

\subsection{Independent Detector Calibration and Tri-Modal Fusion}
Each detector $j\in\{T,K\}$ produces a scalar score $x_{j,i}$, oriented so larger values indicate an anomaly. Detector-specific configuration is chosen independently. For the adaptive fusion variant, let $\mu_j$ denote the calibrated standalone decision threshold and $\sigma_j$ the training-score standard deviation. Define the signed margin
\begin{equation}
 z_{j,i}=\frac{x_{j,i}-\mu_j}{\sigma_j+\epsilon},\qquad j\in\{T,K\}.
\end{equation}
The image-dependent weights and fused prediction are
\begin{equation}
 w_{j,i}=\frac{\exp(|z_{j,i}|/\tau)}{\sum_{\ell\in\{T,F,K\}}\exp(|z_{\ell,i}|/\tau)},\quad
 S_i=\sum_{j\in\{T,F,K\}}w_{j,i}z_{j,i},\quad
 \hat y_i=\mathbb I[S_i\ge d].
 \label{eq:trimodal_fusion}
\end{equation}
The absolute margin sets each detector's relative confidence, whereas its sign determines whether it supports or opposes an anomaly decision. The temperature $\tau$ and final threshold $d$ are calibrated using training data, with no neural-network optimization. The implementation additionally evaluates equal-weight, maximum, AUC-weighted, logistic, and simplex-constrained weighted fusion as comparison variants; these should not be conflated with the adaptive-softmax formulation.

\subsection{Evaluation Protocol and Scope}
Calibration, score orientation, threshold selection, and fusion fitting are restricted to the training data for each split in the intended evaluation protocol. Held-out images are used only for evaluation. The existing biological-cell experiments report NCC--KDE two-detector fusion. A unified tri-modal comparison on the target datasets will be added when the remaining experiments are available.

\section{Experiments and Results}
\subsection{Dataset and Evaluation Protocol}
The earlier manuscript reports 245 positive and 204 negative biological-cell images (449 total), whereas the new experiment log refers to 450 images. We preserve this discrepancy pending confirmation of the exact dataset version and split membership. We report precision, recall, FPR and F1; test FPR below 5\% is the target operating condition. The new training-fraction comparisons do not share a fixed held-out test set, and some ResNet-50 epochs were retrospectively selected using test metrics. Consequently these numbers are descriptive, not a fully controlled model-selection comparison.
\subsection{Qualitative Localization}
\begin{figure}[t]
\centering

\begin{minipage}[t]{0.47\textwidth}
\vspace{0pt}
\centering
\safeimage{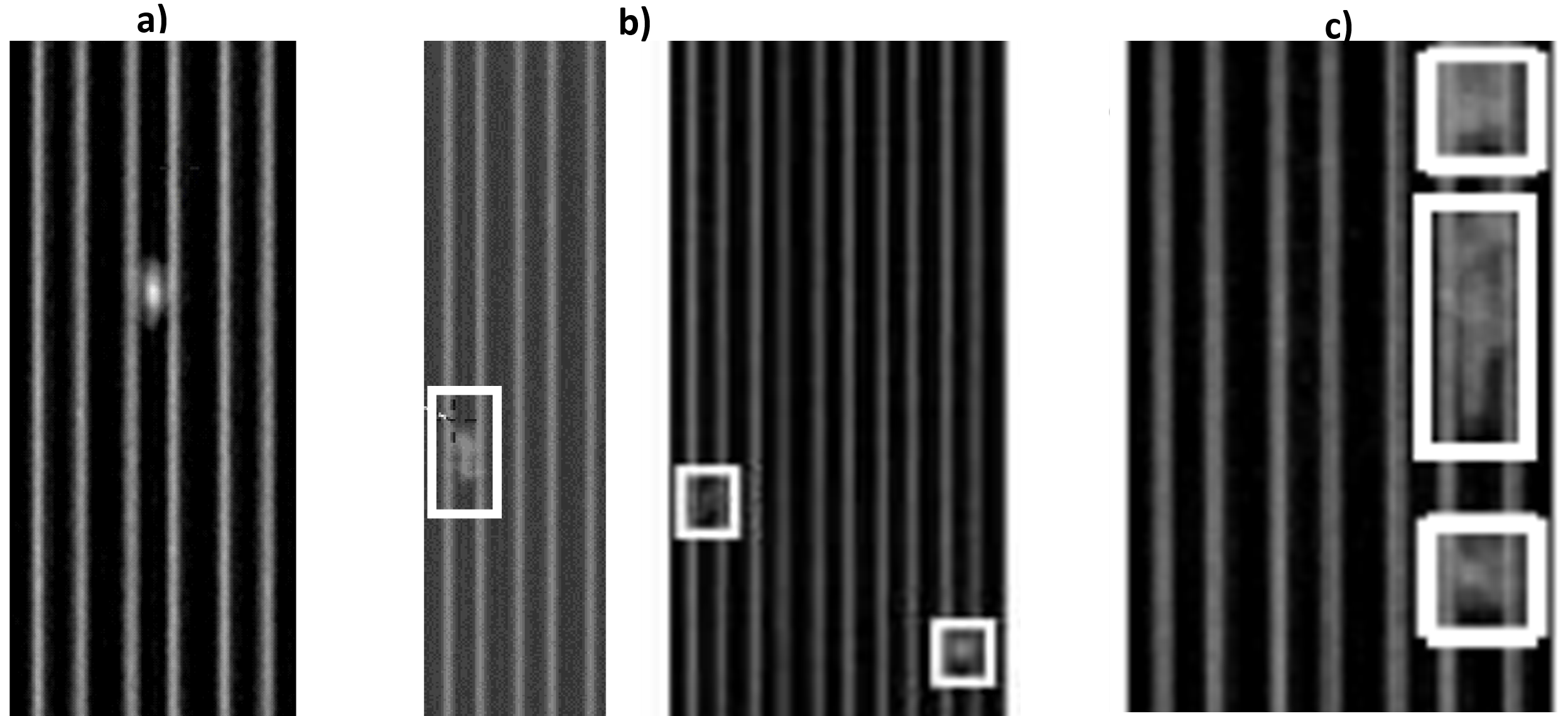}{\linewidth}
\caption{Template matching results:
(a) variational matching handles aspect-ratio changes;
(b) detection of diverse anomaly shapes;
(c) detection of multiple anomalies in one image.}
\label{fig:fig8}
\end{minipage}
\hfill
\begin{minipage}[t]{0.51\textwidth}
\vspace{0pt}
\centering
\captionsetup{type=table}
\captionof{table}{Previously reported baseline results.
The evaluation protocol differs from Table~\ref{tab:all}.}
\label{tab:previous_baselines}
\scriptsize
\setlength{\tabcolsep}{3pt}
\begin{tabular}{lcccc}
\toprule
Method & Recall & FPR & F1 & Loc. \\
\midrule
Fixed template       & .680 & .152 & .755 & Yes \\
Variational template & .820 & .049 & .881 & Yes \\
Original fixed fusion & .882 & .038 & .920 & Yes \\
ResNet-50            & .860 & .037 & .909 & No  \\
\bottomrule
\end{tabular}
\end{minipage}

\end{figure}

Variational matching detects small debris, elongated or irregular breakage, and multiple defects within a single image. Figure~\ref{fig:fig8} illustrates the advantage of transformed templates over a fixed template when anomaly aspect ratios vary, as well as the ability to localize defects with different shapes and orientations. NCC reduces sensitivity to uniform intensity shifts relative to unnormalized correlation; the qualitative examples specifically demonstrate geometric robustness and localization, rather than a controlled illumination ablation.

\subsection{Earlier Baseline Comparison (Original Study)}
The original manuscript reported the following results under its earlier protocol. These numbers are retained for provenance, not pooled with the new training-fraction study; the original description of fusion as training-free does not apply to the newly calibrated adaptive fusion.

\subsection{New Training-Fraction Comparison}

We vary the training fraction of the 450-image dataset and report
precision (P), recall (R), false-positive rate (FPR), and F1 on
each corresponding test partition. The intended operating requirement
is a strict FPR below 5\%. Accordingly, the appropriate model
selection objective is
\begin{equation}
\theta^*=\arg\max_{\theta:\mathrm{FPR}_{\rm val}(\theta)<0.05}
\mathrm{F1}_{\rm val}(\theta),
\label{eq:selection}
\end{equation}
followed by evaluation on held-out test data. The available fusion
sweep was instead selected by training F1 without this FPR constraint,
and the reported CNN checkpoint selection used test metrics. We
therefore retain both passing and failing test results transparently
rather than presenting the existing sweep as a completed constrained
comparison.

\begin{table}[t]
\centering
\small
\caption{Proposed adaptive fusion versus ResNet-50
at different training fractions. Each entry reports
F1/FPR. CNN results include early training
(epochs 1 and 3) and the best reported checkpoint
satisfying FPR $<5\%$ within 20 epochs.
Detailed detector results are given in Appendix~\ref{app:detectors}, and full CNN epoch logs in Appendix~\ref{app:cnn}.}
\label{tab:all}
\resizebox{0.7\columnwidth}{!}{%
\begin{tabular}{lcccc}
\toprule
Train & Fusion & CNN (1 ep.) & CNN (3 ep.)
& CNN (best) \\
\midrule
5\%  & .912/.138 & .118/.118
     & .241/.041 & .349/.041 \\
10\% & .915/.119 & .142/.146
     & .609/.114 & .269/.022 \\
20\% & .938/.061 & .188/.037
     & .845/.256 & .957/.000 \\
30\% & .924/.090 & .747/.181
     & .857/.333 & .967/.007 \\
50\% & .924/.039 & .827/.320
     & .925/.136 & .992/.010 \\
\bottomrule
\end{tabular}}
\end{table}

\paragraph{Early-epoch comparison.}
Table~\ref{tab:all} isolates the effect of CNN training duration by reporting ResNet-50 after one and three epochs alongside its best reported checkpoint satisfying test FPR $<5\%$ within 20 epochs. NCC--KDE adaptive fusion attains higher F1 than the one-epoch CNN at every evaluated training fraction. After three epochs, fusion retains a higher F1 at four of five fractions; at 50\% training, the methods are essentially tied (0.924 versus 0.925). These results characterize performance during early CNN optimization, not a matched-compute comparison: the non-neural detectors incur their own feature-extraction and calibration costs, and the compared operating thresholds have not been optimized identically.

\paragraph{Low-data behavior and false alarms.}
At 5\% and 10\% training, fusion yields F1=0.912 and 0.915, respectively. The corresponding test FPRs, however, are 13.8\% and 11.9\%, exceeding the specified 5\% limit. The CNN's reported FPR-feasible checkpoints yield F1=0.349 and 0.269, but these values should not be compared as though the fusion operating points were equally feasible. The full epoch logs (Appendix~\ref{app:cnn}) additionally show that the CNN can reach higher unconstrained F1 than its FPR-feasible checkpoints. The standalone NCC results across training fractions and the available KDE result are documented in Appendix~\ref{app:detectors}.

\paragraph{Performance with additional CNN training.}
At 20\% training, the reported feasible CNN checkpoint achieves F1=0.957 with no observed false positives, compared with fusion F1=0.938 at FPR=6.1\%. At 30\%, the CNN achieves F1=0.967 and FPR=0.7\%, whereas fusion reaches F1=0.924 and FPR=9.0\%. At 50\%, both methods meet the FPR target, but the CNN performs better (F1=0.992 versus 0.924). Thus the results do not support a general claim that fusion outperforms a sufficiently trained CNN. They instead identify a useful early-training comparison and motivate improved calibration of the fusion decision boundary.

\paragraph{Evaluation limitations.}
Training fractions currently correspond to different test partitions, and the reported CNN ``best'' checkpoints were selected retrospectively using test outcomes. These limitations prevent treating the table as a controlled comparison of model selection or label efficiency. A stronger protocol would use a fixed untouched test set, validation-only threshold and checkpoint selection, and multiple stratified training subsets. In particular, fusion should be re-evaluated after optimizing its threshold under the FPR constraint in Eq.~\eqref{eq:selection}; an unconstrained F1 advantage cannot establish superiority at a constrained operating point.

\subsection{Additional Experiments Reserved for Next Revision}
\noindent\textit{Planned insertion points (results not yet provided):} (a) controlled FPR-constrained fusion sweeps on a common held-out test set; (b) adaptive-versus-fixed fusion ablation; (c) semivariogram spatial-filter ablation and sensitivity to KDE bandwidth; (d) score distributions and threshold/decision-boundary visualizations; and (e) computational cost and localization analysis. These are editorial placeholders, not claims that experiments have been completed.
\section{Discussion}
The two detectors target different manifestations of anomalies. Variational NCC responds to recognizable structures despite changes in geometric form; semivariogram-guided KDE combines spatial irregularity screening with intensity-distribution analysis. This complementarity motivates fusion, but it does not guarantee improved performance: an unusually large signed margin can reflect a confident error, and a softmax based on absolute margin magnitude may amplify that error. Calibration quality and the interaction between detector scores therefore matter as much as the availability of multiple signals.

The low-data results reveal a practical tension between sensitivity and false alarms. Fusion maintains relatively high unconstrained F1 with small training fractions, yet most of those configurations violate the target FPR. Conversely, ResNet-50 achieves substantially stronger feasible performance after sufficient training. Future evaluation should therefore emphasize F1 at a prespecified FPR limit, the stability of calibration across splits, and the total cost of feature extraction and model optimization, rather than F1 or epoch count alone.

\section{Conclusion}
We present a detection framework combining variational geometric matching and semivariogram-guided intensity analysis through independently calibrated, image-adaptive fusion. The current biological-image experiments support the geometric and NCC--KDE components and show that calibrated fusion can be competitive with ResNet-50 early in training. They also show that a fully trained CNN achieves higher accuracy and that several fusion operating points fail a strict 5\% FPR requirement.

\appendix
\renewcommand{\thesection}{Appendix \Alph{section}}

\section{Detailed Detector Results for Patterned Images}
\label{app:detectors}
\label{app:detectors}
\begin{table}[t]
\centering
\small
\caption{Detailed standalone detector and NCC--KDE fusion
results across training fractions. Asterisked F1 values
violate the strict test-FPR $<5\%$ criterion.
KDE has one reported result without a specified
training fraction. Different fractions use different
test partitions.}
\label{tab:appendix_detectors}
\resizebox{0.7\columnwidth}{!}{%
\begin{tabular}{llrrrrl}
\toprule
Method & Train (\%) & P & R & FPR & F1 & Status \\
\midrule
NCC template & 0 (fixed $d=.50$)
 & .947 & .878 & .059 & .911$^*$ & Fail \\
 & 10 & .989 & .801 & .011 & .885 & Pass \\
 & 20 & .987 & .791 & .012 & .878 & Pass \\
 & 30 & .968 & .878 & .035 & .921 & Pass \\
 & 40 & .963 & .884 & .041 & .922 & Pass \\
 & 50 & .956 & .878 & .049 & .915 & Pass \\
 & 60 & .978 & .898 & .024 & .936 & Pass \\
 & 70 & .985 & .892 & .016 & .936 & Pass \\
\midrule
Semivariogram--KDE & Not specified
 & .967 & .848 & .035 & .903 & Pass \\
\midrule
Adaptive NCC--KDE fusion & 5
 & .890 & .936 & .138 & .912$^*$ & Fail \\
 & 10 & .903 & .928 & .119 & .915$^*$ & Fail \\
 & 20 & .948 & .929 & .061 & .938$^*$ & Fail \\
 & 30 & .924 & .924 & .090 & .924$^*$ & Fail \\
 & 50 & .965 & .886 & .039 & .924 & Pass \\
\bottomrule
\end{tabular}}
\end{table}

\section{Complete ResNet-50 Epoch-Level Results}
\label{app:cnn}
\label{app:cnn_epochs}
Tables~\ref{tab:cnn5}--\ref{tab:cnn50} reproduce all 20 epochs for each training fraction from the experimental logs. A row is feasible only when test FPR $<0.05$. Bold F1 marks the highest F1 among feasible epochs in that split. This is a retrospective test-based selection for reporting; future model selection should use a validation set.
\begin{table}[htbp]
\centering
\caption{ResNet-50 epoch results: 5\% training ($n_{\mathrm{train}}=22$, $n_{\mathrm{test}}=428$).}
\label{tab:cnn5}
\small
\begin{tabular}{rrrrrrrrl}
\toprule
Epoch & Train loss & Train F1 & Test loss & P & R & FPR & F1 & Feasible \\
\midrule
1 & 0.7085 & 0.300 & 0.7021 & 0.410 & 0.069 & 0.118 & 0.118 & No \\
2 & 0.5979 & 0.917 & 0.6927 & 0.562 & 0.077 & 0.072 & 0.136 & No \\
3 & 0.5438 & 0.833 & 0.6798 & 0.805 & 0.142 & 0.041 & 0.241 & Yes \\
4 & 0.4799 & 0.923 & 0.6716 & 0.864 & 0.219 & 0.041 & \textbf{0.349} & Yes \\
5 & 0.4367 & 0.960 & 0.6650 & 0.855 & 0.305 & 0.062 & 0.449 & No \\
6 & 0.3619 & 1.000 & 0.6586 & 0.857 & 0.412 & 0.082 & 0.557 & No \\
7 & 0.3328 & 0.957 & 0.6512 & 0.800 & 0.515 & 0.154 & 0.627 & No \\
8 & 0.3292 & 0.957 & 0.6408 & 0.783 & 0.605 & 0.200 & 0.683 & No \\
9 & 0.2345 & 1.000 & 0.6293 & 0.763 & 0.678 & 0.251 & 0.718 & No \\
10 & 0.1993 & 0.960 & 0.6187 & 0.753 & 0.721 & 0.282 & 0.737 & No \\
11 & 0.1477 & 1.000 & 0.6089 & 0.744 & 0.773 & 0.318 & 0.758 & No \\
12 & 0.1321 & 1.000 & 0.5921 & 0.749 & 0.833 & 0.333 & 0.789 & No \\
13 & 0.1158 & 1.000 & 0.5786 & 0.756 & 0.850 & 0.328 & 0.800 & No \\
14 & 0.0938 & 1.000 & 0.5569 & 0.763 & 0.858 & 0.318 & 0.808 & No \\
15 & 0.0815 & 1.000 & 0.5441 & 0.777 & 0.880 & 0.303 & 0.825 & No \\
16 & 0.1124 & 1.000 & 0.5254 & 0.781 & 0.888 & 0.297 & 0.831 & No \\
17 & 0.0570 & 1.000 & 0.5142 & 0.779 & 0.863 & 0.292 & 0.819 & No \\
18 & 0.0451 & 1.000 & 0.5000 & 0.788 & 0.863 & 0.277 & 0.824 & No \\
19 & 0.0748 & 1.000 & 0.4846 & 0.787 & 0.871 & 0.282 & 0.827 & No \\
20 & 0.0487 & 1.000 & 0.4755 & 0.786 & 0.803 & 0.262 & 0.794 & No \\
\bottomrule
\end{tabular}
\end{table}

\begin{table}[htbp]
\centering
\caption{ResNet-50 epoch results: 10\% training ($n_{\mathrm{train}}=44$, $n_{\mathrm{test}}=406$).}
\label{tab:cnn10}
\small
\begin{tabular}{rrrrrrrrl}
\toprule
Epoch & Train loss & Train F1 & Test loss & P & R & FPR & F1 & Feasible \\
\midrule
1 & 0.6851 & 0.565 & 0.6993 & 0.413 & 0.086 & 0.146 & 0.142 & No \\
2 & 0.5687 & 0.800 & 0.6782 & 0.897 & 0.158 & 0.022 & \textbf{0.269} & Yes \\
3 & 0.4554 & 0.980 & 0.6595 & 0.835 & 0.480 & 0.114 & 0.609 & No \\
4 & 0.3803 & 0.960 & 0.6426 & 0.794 & 0.647 & 0.200 & 0.713 & No \\
5 & 0.2659 & 1.000 & 0.6230 & 0.761 & 0.805 & 0.303 & 0.782 & No \\
6 & 0.2342 & 0.960 & 0.6025 & 0.763 & 0.887 & 0.330 & 0.820 & No \\
7 & 0.2144 & 0.960 & 0.5783 & 0.787 & 0.905 & 0.292 & 0.842 & No \\
8 & 0.1468 & 1.000 & 0.5493 & 0.809 & 0.919 & 0.259 & 0.860 & No \\
9 & 0.1012 & 1.000 & 0.5143 & 0.816 & 0.882 & 0.238 & 0.848 & No \\
10 & 0.0754 & 1.000 & 0.4861 & 0.843 & 0.878 & 0.195 & 0.860 & No \\
11 & 0.0509 & 1.000 & 0.4486 & 0.849 & 0.864 & 0.184 & 0.857 & No \\
12 & 0.0405 & 1.000 & 0.4218 & 0.867 & 0.824 & 0.151 & 0.845 & No \\
13 & 0.0249 & 1.000 & 0.4025 & 0.879 & 0.792 & 0.130 & 0.833 & No \\
14 & 0.0294 & 1.000 & 0.3881 & 0.881 & 0.769 & 0.124 & 0.821 & No \\
15 & 0.0298 & 1.000 & 0.3915 & 0.881 & 0.769 & 0.124 & 0.821 & No \\
16 & 0.0172 & 1.000 & 0.3860 & 0.886 & 0.774 & 0.119 & 0.826 & No \\
17 & 0.0193 & 1.000 & 0.3857 & 0.903 & 0.756 & 0.097 & 0.823 & No \\
18 & 0.0122 & 1.000 & 0.3821 & 0.894 & 0.765 & 0.108 & 0.824 & No \\
19 & 0.0129 & 1.000 & 0.3925 & 0.909 & 0.769 & 0.092 & 0.833 & No \\
20 & 0.0309 & 1.000 & 0.3845 & 0.908 & 0.760 & 0.092 & 0.828 & No \\
\bottomrule
\end{tabular}
\end{table}

\begin{table}[htbp]
\centering
\caption{ResNet-50 epoch results: 20\% training ($n_{\mathrm{train}}=90$, $n_{\mathrm{test}}=360$).}
\label{tab:cnn20}
\small
\begin{tabular}{rrrrrrrrl}
\toprule
Epoch & Train loss & Train F1 & Test loss & P & R & FPR & F1 & Feasible \\
\midrule
1 & 0.6764 & 0.533 & 0.6818 & 0.778 & 0.107 & 0.037 & 0.188 & Yes \\
2 & 0.5104 & 0.883 & 0.6425 & 0.845 & 0.668 & 0.146 & 0.746 & No \\
3 & 0.3917 & 0.891 & 0.6009 & 0.806 & 0.888 & 0.256 & 0.845 & No \\
4 & 0.2872 & 0.942 & 0.5671 & 0.808 & 0.944 & 0.268 & 0.871 & No \\
5 & 0.2165 & 0.950 & 0.5111 & 0.831 & 0.954 & 0.232 & 0.888 & No \\
6 & 0.1465 & 0.970 & 0.4278 & 0.885 & 0.939 & 0.146 & 0.911 & No \\
7 & 0.1587 & 0.948 & 0.3632 & 0.944 & 0.949 & 0.067 & 0.947 & No \\
8 & 0.0971 & 0.990 & 0.2880 & 0.973 & 0.923 & 0.030 & 0.948 & Yes \\
9 & 0.0748 & 1.000 & 0.2299 & 0.984 & 0.913 & 0.018 & 0.947 & Yes \\
10 & 0.0497 & 1.000 & 0.2048 & 0.989 & 0.903 & 0.012 & 0.944 & Yes \\
11 & 0.0290 & 1.000 & 0.1949 & 0.989 & 0.878 & 0.012 & 0.930 & Yes \\
12 & 0.0175 & 1.000 & 0.1881 & 0.988 & 0.867 & 0.012 & 0.924 & Yes \\
13 & 0.0200 & 1.000 & 0.1858 & 0.983 & 0.872 & 0.018 & 0.924 & Yes \\
14 & 0.0184 & 1.000 & 0.1937 & 0.977 & 0.862 & 0.024 & 0.916 & Yes \\
15 & 0.0206 & 0.990 & 0.2041 & 0.982 & 0.847 & 0.018 & 0.910 & Yes \\
16 & 0.0091 & 1.000 & 0.2236 & 0.994 & 0.827 & 0.006 & 0.903 & Yes \\
17 & 0.0071 & 1.000 & 0.2014 & 1.000 & 0.832 & 0.000 & 0.908 & Yes \\
18 & 0.0039 & 1.000 & 0.1775 & 1.000 & 0.852 & 0.000 & 0.920 & Yes \\
19 & 0.0033 & 1.000 & 0.1497 & 1.000 & 0.878 & 0.000 & 0.935 & Yes \\
20 & 0.0273 & 1.000 & 0.1251 & 1.000 & 0.918 & 0.000 & \textbf{0.957} & Yes \\
\bottomrule
\end{tabular}
\end{table}

\begin{table}[htbp]
\centering
\caption{ResNet-50 epoch results: 30\% training ($n_{\mathrm{train}}=134$, $n_{\mathrm{test}}=316$).}
\label{tab:cnn30}
\small
\begin{tabular}{rrrrrrrrl}
\toprule
Epoch & Train loss & Train F1 & Test loss & P & R & FPR & F1 & Feasible \\
\midrule
1 & 0.6389 & 0.732 & 0.6424 & 0.819 & 0.686 & 0.181 & 0.747 & No \\
2 & 0.4606 & 0.903 & 0.6032 & 0.742 & 0.855 & 0.354 & 0.795 & No \\
3 & 0.3137 & 0.899 & 0.5199 & 0.775 & 0.959 & 0.333 & 0.857 & No \\
4 & 0.2030 & 0.960 & 0.4013 & 0.786 & 0.959 & 0.312 & 0.864 & No \\
5 & 0.1451 & 0.986 & 0.3119 & 0.825 & 0.930 & 0.236 & 0.874 & No \\
6 & 0.1011 & 0.993 & 0.2576 & 0.860 & 0.930 & 0.181 & 0.894 & No \\
7 & 0.0632 & 1.000 & 0.1955 & 0.906 & 0.953 & 0.118 & 0.929 & No \\
8 & 0.0408 & 0.993 & 0.1725 & 0.947 & 0.930 & 0.062 & 0.938 & No \\
9 & 0.0285 & 1.000 & 0.1455 & 0.958 & 0.930 & 0.049 & 0.944 & Yes \\
10 & 0.0171 & 1.000 & 0.1322 & 0.970 & 0.948 & 0.035 & 0.959 & Yes \\
11 & 0.0157 & 1.000 & 0.1311 & 0.987 & 0.913 & 0.014 & 0.949 & Yes \\
12 & 0.0143 & 1.000 & 0.1269 & 0.981 & 0.924 & 0.021 & 0.952 & Yes \\
13 & 0.0099 & 1.000 & 0.1400 & 0.970 & 0.924 & 0.035 & 0.946 & Yes \\
14 & 0.0171 & 0.993 & 0.1257 & 0.988 & 0.930 & 0.014 & 0.958 & Yes \\
15 & 0.0108 & 0.993 & 0.1313 & 0.988 & 0.930 & 0.014 & 0.958 & Yes \\
16 & 0.0201 & 0.993 & 0.1437 & 0.982 & 0.936 & 0.021 & 0.958 & Yes \\
17 & 0.0108 & 1.000 & 0.1399 & 0.981 & 0.924 & 0.021 & 0.952 & Yes \\
18 & 0.0117 & 1.000 & 0.1485 & 0.994 & 0.913 & 0.007 & 0.952 & Yes \\
19 & 0.0096 & 1.000 & 0.1454 & 0.994 & 0.942 & 0.007 & \textbf{0.967} & Yes \\
20 & 0.0039 & 1.000 & 0.1291 & 0.988 & 0.942 & 0.014 & 0.964 & Yes \\
\bottomrule
\end{tabular}
\end{table}

\begin{table}[htbp]
\centering
\caption{ResNet-50 epoch results: 50\% training ($n_{\mathrm{train}}=224$, $n_{\mathrm{test}}=226$).}
\label{tab:cnn50}
\small
\begin{tabular}{rrrrrrrrl}
\toprule
Epoch & Train loss & Train F1 & Test loss & P & R & FPR & F1 & Feasible \\
\midrule
1 & 0.6118 & 0.776 & 0.6099 & 0.769 & 0.894 & 0.320 & 0.827 & No \\
2 & 0.3832 & 0.897 & 0.5112 & 0.821 & 0.967 & 0.252 & 0.888 & No \\
3 & 0.2298 & 0.947 & 0.3032 & 0.894 & 0.959 & 0.136 & 0.925 & No \\
4 & 0.1117 & 0.988 & 0.1786 & 0.967 & 0.943 & 0.039 & 0.955 & Yes \\
5 & 0.0689 & 0.992 & 0.1157 & 0.983 & 0.967 & 0.019 & 0.975 & Yes \\
6 & 0.0454 & 0.996 & 0.0910 & 0.992 & 0.967 & 0.010 & 0.979 & Yes \\
7 & 0.0283 & 1.000 & 0.0816 & 0.992 & 0.976 & 0.010 & 0.984 & Yes \\
8 & 0.0207 & 1.000 & 0.0604 & 0.992 & 0.976 & 0.010 & 0.984 & Yes \\
9 & 0.0150 & 1.000 & 0.0477 & 1.000 & 0.976 & 0.000 & 0.988 & Yes \\
10 & 0.0084 & 1.000 & 0.0492 & 1.000 & 0.959 & 0.000 & 0.979 & Yes \\
11 & 0.0058 & 1.000 & 0.0475 & 1.000 & 0.967 & 0.000 & 0.983 & Yes \\
12 & 0.0054 & 1.000 & 0.0445 & 1.000 & 0.967 & 0.000 & 0.983 & Yes \\
13 & 0.0026 & 1.000 & 0.0470 & 1.000 & 0.967 & 0.000 & 0.983 & Yes \\
14 & 0.0033 & 1.000 & 0.0398 & 1.000 & 0.984 & 0.000 & \textbf{0.992} & Yes \\
15 & 0.0029 & 1.000 & 0.0417 & 1.000 & 0.976 & 0.000 & 0.988 & Yes \\
16 & 0.0032 & 1.000 & 0.0321 & 1.000 & 0.984 & 0.000 & 0.992 & Yes \\
17 & 0.0021 & 1.000 & 0.0297 & 1.000 & 0.984 & 0.000 & 0.992 & Yes \\
18 & 0.0012 & 1.000 & 0.0323 & 1.000 & 0.984 & 0.000 & 0.992 & Yes \\
19 & 0.0035 & 1.000 & 0.0363 & 0.992 & 0.992 & 0.010 & 0.992 & Yes \\
20 & 0.0022 & 1.000 & 0.0333 & 0.992 & 0.992 & 0.010 & 0.992 & Yes \\
\bottomrule
\end{tabular}
\end{table}


\begin{thebibliography}{99}
\bibitem{Aksoy2004}
Aksoy, M. S., Torkul, O., and Cedimoglu, I. H. (2004).
An industrial visual inspection system using inductive learning.
\textit{Journal of Intelligent Manufacturing}, 15(4), 569--574.

\bibitem{Bergmann2019}
Bergmann, P., Fauser, M., Sattlegger, D., and Steger, C. (2019).
MVTec AD: A comprehensive real-world dataset for unsupervised anomaly detection.
In \textit{Proceedings of the IEEE/CVF Conference on Computer Vision and Pattern Recognition (CVPR)}, 9592--9600.

\bibitem{Briechle2001}
Briechle, K., and Hanebeck, U. D. (2001).
Template matching using fast normalized cross correlation.
In \textit{Proceedings of SPIE}, 4387, 95--102.

\bibitem{Brunelli2009}
Brunelli, R. (2009).
\textit{Template Matching Techniques in Computer Vision: Theory and Practice}.
Wiley.

\bibitem{Defard2021}
Defard, T., Setkov, A., Loesch, A., and Audigier, R. (2021).
PaDiM: A patch distribution modeling framework for anomaly detection and localization.
In \textit{Proceedings of the International Conference on Pattern Recognition (ICPR)}, 475--489.

\bibitem{He2016}
He, K., Zhang, X., Ren, S., and Sun, J. (2016).
Deep residual learning for image recognition.
In \textit{Proceedings of the IEEE Conference on Computer Vision and Pattern Recognition (CVPR)}, 770--778.

\bibitem{Lewis1995}
Lewis, J. P. (1995).
Fast normalized cross-correlation.
In \textit{Vision Interface}, 120--123.

\bibitem{Li2021}
Li, C.-L., Sohn, K., Yoon, J., and Pfister, T. (2021).
CutPaste: Self-supervised learning for anomaly detection and localization.
In \textit{Proceedings of the IEEE/CVF Conference on Computer Vision and Pattern Recognition (CVPR)}, 9664--9674.

\bibitem{Roth2022}
Roth, K., Pemula, L., Zepeda, J., Schölkopf, B., Brox, T., and Gehler, P. (2022).
Towards total recall in industrial anomaly detection.
In \textit{Proceedings of the IEEE/CVF Conference on Computer Vision and Pattern Recognition (CVPR)}, 14318--14328.

\bibitem{Szeliski2022}
Szeliski, R. (2022).
\textit{Computer Vision: Algorithms and Applications}, 2nd ed.
Springer.

\bibitem{Tahmasebi2012}
Tahmasebi, P., Hezarkhani, A., and Sahimi, M. (2012).
Multiple-point geostatistical modeling based on cross-correlation functions.
\textit{Computational Geosciences}, 16(3), 779--797.
\bibitem{Parzen1962}
Parzen, E. (1962).
On estimation of a probability density function.
\textit{Annals of Mathematical Statistics}.

\bibitem{Silverman1986}
Silverman, B. (1986).
\textit{Density Estimation for Statistics and Data Analysis}.
Chapman and Hall.
\end{thebibliography}
\end{document}